%% file: acl2023.tex
\pdfoutput=1

\documentclass[11pt]{article}

\usepackage[]{ACL2023}

\usepackage{times}
\usepackage{latexsym}

\usepackage[T1]{fontenc}

\usepackage[utf8]{inputenc}

\usepackage{microtype}

\usepackage{inconsolata}
\usepackage{multicol}
\usepackage{booktabs}
\usepackage{colortbl}
\usepackage[fleqn]{amsmath}
\usepackage{multirow}
\usepackage[normalem]{ulem}
\useunder{\uline}{\ul}{}
\usepackage{hyperref}
\usepackage{graphicx}
\usepackage{subfigure}
\usepackage{fixltx2e}
\usepackage{xcolor}
\usepackage{algorithm}
\usepackage{algorithmic}
\usepackage{setspace}
\usepackage{amsfonts}
\usepackage[most]{tcolorbox}
\usepackage{afterpage}

\newcommand{\ie}{\textit{i.e.}}
\newcommand{\eg}{\textit{e.g.}}
\newcommand{\ours}{\texttt{mmE5}}
\newcommand{\todo}[1]{\{\textcolor{blue}{\textbf{TODO}}\}}

\title{\ours{}: Improving Multimodal Multilingual Embeddings via \\ High-quality Synthetic Data}

\author{Haonan Chen$^{1}$\thanks{$^{*}$Work done during Haonan’s internship at MSR Asia. Prof. Zhicheng Dou is the corresponding author.}, Liang Wang$^2$, Nan Yang$^2$, Yutao Zhu$^1$ \\ \textbf{Ziliang Zhao$^1$, Furu Wei$^2$, Zhicheng Dou$^{1}$} \\
        $^1$Gaoling School of Artificial Intelligence, Renmin University of China \\ 
        $^2$Microsoft Corporation \\ 
        \texttt{\{hnchen,dou\}@ruc.edu.cn} \\
        \texttt{\{wangliang,nanya,fuwei\}@microsoft.com} \\
        \url{https://github.com/haon-chen/mmE5} \\
}

\newtcolorbox[list inside=prompt]{prompt}[1][]{
    colbacktitle=black!60,
    coltitle=white,
    fontupper=\footnotesize,
    boxsep=5pt,
    left=0pt,
    right=-1pt,
    top=0pt,
    bottom=0pt,
    boxrule=1pt,
    width=\textwidth,
    #1,
}
\begin{document}
\maketitle

\begin{abstract}

Multimodal embedding models have gained significant attention for their ability to map data from different modalities, such as text and images, into a unified representation space. 
However, the limited labeled multimodal data often hinders embedding performance. 
Recent approaches have leveraged data synthesis to address this problem, yet the quality of synthetic data remains a critical bottleneck. 
In this work, we identify three criteria for high-quality synthetic multimodal data. First, \textbf{broad scope} ensures that the generated data covers diverse tasks and modalities, making it applicable to various downstream scenarios. 
Second, \textbf{robust cross-modal alignment} makes different modalities semantically consistent. 
Third, \textbf{high fidelity} ensures that the synthetic data maintains realistic details to enhance its reliability.
Guided by these principles, we synthesize datasets that: (1) cover a wide range of tasks, modality combinations, and languages, (2) are generated via a deep thinking process within a single pass of a multimodal large language model, and (3) incorporate real-world images with accurate and relevant texts, ensuring fidelity through self-evaluation and refinement.
Leveraging these high-quality synthetic and labeled datasets, we train a \textbf{m}ultimodal \textbf{m}ultilingual \textbf{E5} model \ours{}. 
Extensive experiments demonstrate that \ours{} achieves state-of-the-art performance on the MMEB Benchmark and superior multilingual performance on the XTD benchmark.
Our codes, datasets and models are released in \url{https://github.com/haon-chen/mmE5}.

\end{abstract}
\input{sections/1_introduction.tex}

\input{sections/2_related_work.tex}
\input{sections/3_methodology.tex}

\input{sections/4_experiments.tex}

\input{sections/5_conclusion.tex}

\input{acl2023.bbl}
\appendix

\input{sections/appendix.tex}

\end{document}

%% file: sections/1_introduction.tex
\section{Introduction}

Multimodal embedding models encode multimedia inputs, such as images and text, into latent vector representations.
They have demonstrated effectiveness across diverse downstream tasks, including classification~\cite{ImageNet-1K}, visual question answering (VQA)~\cite{TextVQA}, and cross-modal retrieval~\cite{OVEN}. 
Prior studies have focused on training multimodal embedding models using simple text-image pre-trained models such as CLIP~\cite{CLIP}. 
More recently, researchers have turned to multimodal large language models (MLLMs), including LLaVA~\cite{Llava} and Phi~\cite{Phi3}, to develop universal embedding models.

These vision-language models (VLMs) mostly rely on high-quality human-labeled datasets to achieve robust embedding capabilities. 
Such datasets suffer from data scarcity because they require high costs of multimodal annotations.
To address this, researchers have leveraged the advanced language modeling capabilities of large language models (LLMs) and MLLMs to synthesize datasets for fine-tuning multimodal embedding models~\cite{MagicLens, megapairs, GME}.
However, existing works lack a comprehensive exploration into the quality of synthetic embedding data.
Typically, most data generated by them are limited to specific modality types of English retrieval tasks, harming the generalization capabilities of the embedding models.

\begin{figure}[!t]
	\centering
	\includegraphics[width=1.0\linewidth]{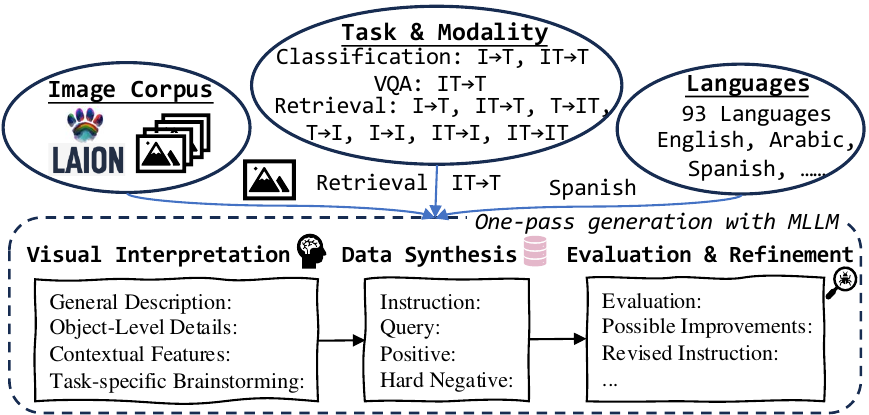}
	\caption{An illustration of our data synthesis framework. ``X$\rightarrow$Y'' denotes a modality combination, where ``X'' represents the query side and ``Y'' denotes the target side. ``T'' denotes text and ``I'' denotes image.}
	\vspace{-2ex}
	\label{fig:introduction}
\end{figure}

After analyzing common application scenarios of multimodal embedding models, we identify three key criteria and introduce a data synthesis framework guided by these principles:
\textbf{(1) Broad scope.}
Multimodal embedding models are commonly employed in tasks such as classification, visual question answering (VQA), and retrieval, which requires understanding various input combinations of text and images.
Additionally, multilingual contexts are increasingly popular in daily scenarios.
As shown in Figure~\ref{fig:introduction}, our framework synthesizes datasets covering three tasks, seven modality combinations, and 93 languages, ensuring that models trained on it generalize effectively across diverse scenarios.
\textbf{(2) Robust cross-modal alignment.}
In multimodal tasks, models must understand and align information across different modalities to generate meaningful representations. 
Without accurate cross-modal alignment, embeddings may fail to capture the underlying relationships, leading to poor performance in downstream tasks.
To synthesize data of robust cross-modal alignment, our framework incorporates a deep thinking process. 
Specifically, for each sampled image, we first employ an MLLM to interpret it from four perspectives before generating data: general information, object-level description, contextual background information, and task-specific brainstorming, \ie, how the image relates to the given task. 
Additionally, the entire data synthesis process is executed within a single pass of an MLLM.
By this, the MLLM can ``see'' the images at the whole time, avoiding potential information loss that might occur due to multiple I/O steps in previous works~\cite{megapairs, GME}.
\textbf{(3) High fidelity.}
The individual quality of each modality (\eg, real images, high-quality instructions, queries, and hard negatives) determines the overall usefulness of the dataset. 
To enhance fidelity, our framework uses real images sampled from an open-source corpus (LAION-400m~\cite{LAION}) as the input images.
We also apply a series of quality control measures, such as self-evaluation and refinement, ensuring that the synthetic components accurately reflect real-world distributions and maintain strong cross-modal alignment.

With the synthesized data ready, we train a \textbf{m}ultimodal \textbf{m}ultilingual \textbf{E5} model (\ours{}).
It achieves state-of-the-art performance on the 36 datasets of MMEB~\cite{MMEB}, using 45 times less training data than the previous SOTA model MMRet~\cite{megapairs} (560K compared to 26M) in a zero-shot setting. 
After incorporating labeled data, \ours{} still demonstrates the best performance.
Besides, \ours{} achieves the best results on the multilingual benchmark XTD~\cite{XTD}, demonstrating its superior multilingual capabilities.

In summary, our contributions are as follows:
\begin{itemize}
    \item Based on our analysis of common scenarios for multimodal embedding models, we identify three key criteria of high-quality synthetic data: broad scope, robust cross-modal alignment, and high fidelity.
    \item We introduce a data synthesis framework guided by the proposed principles. 
    This framework leverages an MLLM to produce high-quality synthetic datasets that cover a wide range of tasks, modality combinations, and languages. It ensures robust cross-modal alignment through a comprehensive multi-aspect interpretation process and maintains high fidelity by employing self-evaluation and refinement mechanisms.
    \item Compared to the previous leading model, \ours{} achieves SOTA performance on the MMEB benchmark while using 45× less synthetic data in both zero-shot and supervised settings.
    \ours{} also demonstrates superior multilingual capabilities on the XTD benchmark.
\end{itemize}

%% file: sections/2_related_work.tex
\section{Related Work}

\noindent \textbf{Multimodal Embedding}
Previous studies, such as CLIP~\cite{CLIP}, Align~\cite{ALIGN}, BLIP~\cite{BLIP}, and CoCa~\cite{CoCa}, have employed large-scale weakly supervised data to learn separate multimodal representations through pre-training.
Some works attempt to obtain universal embeddings for texts and images utilizing existing CLIP-like models~\cite{UniIR, UniVL-DR, VISTA, MARVEL}.
For instance, UniIR~\cite{UniIR} integrates separate embeddings from different modalities into unified features.
Recent approaches finetune MLLMs to leverage their multimodal reasoning capabilities for obtaining universal representations~\cite{E5-V, MMEB, GME, megapairs, mmembed}.
For example, VLM2Vec~\cite{MMEB} utilizes instruction-tuning to transform MLLMs into embedding models.

\begin{table*}[!t]
\centering
\small
\setlength{\tabcolsep}{2.5pt}
\renewcommand{\arraystretch}{1.5}
    \begin{tabular}{l|c|c|l|c|c|c}
    \toprule
Method    & \# Languages & Task                                                                      & Modality Combinations                                                            & w/ MLLM & \multicolumn{1}{l|}{One Pass}  & Self-evaluation \\ \midrule
MagicLens & 1 (English)  & Retrieval                                                                 & IT→I                                                                              & $\times$        & $\surd$                            & $\times$           \\
\hline
MegaPairs     & 1 (English)  & Retrieval                                                                 & IT→I                                                                              & $\surd$        & $\times$                             & $\times$               \\
\hline
GME       & 1 (English)  & Retrieval                                                                 & T→IT, IT→IT                                                                       & $\times$        & $\times$                            & $\times$      \\
\midrule
\ours{} (Ours)      & \begin{tabular}[c]{@{}c@{}}93 (English, \\ Spanish, etc.)\end{tabular}      & \begin{tabular}[c]{@{}c@{}}Classification, \\ VQA, Retrieval\end{tabular} & \begin{tabular}[l]{@{}l@{}}IT→I, T→IT, IT→IT, \\ I→I, I→T, IT→T, T→I\end{tabular} & $\surd$        & $\surd$                             & $\surd$              \\
\bottomrule
    \end{tabular}
    \caption{Comparison of the synthetic datasets in our work with those from previous methods.
    Our synthetic datasets incorporate 93 languages, two additional tasks, and more modality combinations.
    ``IT$\rightarrow$T'' denotes a modality combination, where ``IT'' denotes images and texts on the query side and ``T'' denotes texts on the target side.
    The entire data synthesis process is executed within a single pass of an MLLM, thereby avoiding potential information loss and ensuring robust cross-modal alignment.
    We also employ real images and self-evaluation to maintain fidelity.}
    \label{tab:comparison_syndata}
\end{table*}

\noindent \textbf{Synthetic Data} The generation of synthetic data has been extensively explored for text embedding tasks~\cite{E5mistral, speed, glan}.
With the recent emergence of MLLMs like Phi-3.5-V~\cite{Phi3} and LLaVA~\cite{Llava}, along with diffusion models like Stable Diffusion~\cite{diffusion}, researchers have been focused on synthesizing data to address the scarcity of multimodal instruction-tuning datasets.
For example, MagicLens~\cite{MagicLens} utilizes co-existing images from the same webpage and an LLM to create multimodal data triplets (query image, instruction, relevant image), \ie, IT$\rightarrow$I paradigm.
MegaPairs~\cite{megapairs} aims to synthesize more diverse data triplets by retrieving relevant images from different perspectives.
GME~\cite{GME} employs an LLM and a diffusion model to generate a fused modality dataset that includes both T$\rightarrow$IT and IT$\rightarrow$IT types.
Table~\ref{tab:comparison_syndata} presents a comparison of the synthesized data in this study with that of previous works.

%% file: sections/3_methodology.tex
\section{Methodology: \ours{}}

\begin{figure*}[!t]
	\centering
	\includegraphics[width=1.0\textwidth]{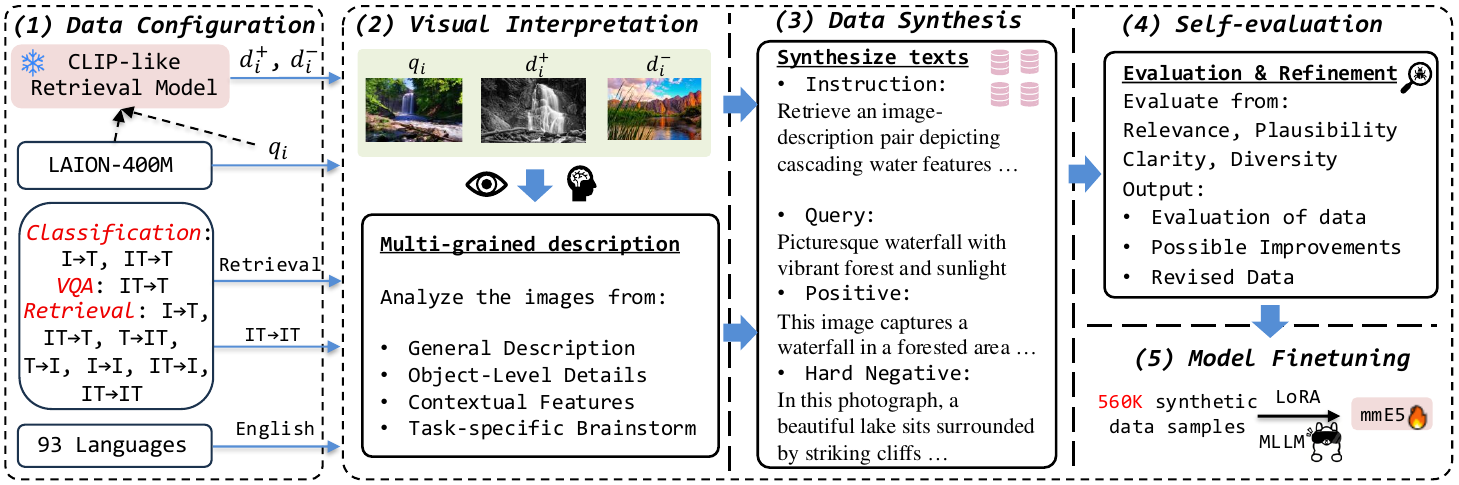}
	\caption{An illustration of our method. We take the generation of an IT$\rightarrow$IT retrieval data sample as an example.}
	\vspace{-2ex}
	\label{fig:framework}
\end{figure*}

In this section, we present our method, which synthesizes high-quality multimodal data for the further finetuning of our embedding model \ours{}.
As shown in Figure~\ref{fig:framework}, our method consists of five stages: 
(1) Initially, for each data sample to be synthesized, we configure the specifics of the task, modality combination, language, and input images.
(2) We employ an MLLM to generate multi-grained descriptions for the input images, ensuring that the synthesized texts are well-aligned with the images.
(3) Utilizing this MLLM, we synthesize text data based on both the images and their descriptions.
(4) The MLLM then evaluates its synthesized data from multiple perspectives, offering revised data to enhance cross-modal alignment and fidelity.
(5) Finally, the synthesized texts and images are used to finetune an MLLM specifically for embedding tasks.
To minimize potential information loss, stages (2), (3), and (4) are executed within a single pass of the MLLM.

\subsection{Preliminaries}

An MLLM can accept text, image, or text-image pairs as input, allowing both the query side $q$ and the document side $d$ to be multimodal.
Inspired by existing works on synthetic text embedding data~\cite{E5mistral, speed}, each data sample we generate is a quadruple of (task instruction, query, positive document, hard negative document), denoted as $(t, q, d^+, d^-)$.
For each data piece, we first sample images from the large-scale open-source image corpus LAION-400M~\cite{LAION} as the query image, positive image, and hard negative image ($q_i$, $d^+_i$, $d^-_i$).
Then, with these three images as input, an MLLM $\pi_{\theta}$ can synthesize a multimodal embedding data sample $ y \sim \pi_{\theta} (y \mid q_i, d^+_i, d^-_i) $, where $y = (t, q_t, d^+_t, d^-_t)$.
As a result, the synthetic data can have a maximum of seven elements: $\{t, (q_t, q_i), (d^+_t, d^+_i), (d^-_t, d^-_i)\}$.
More data examples can be found in Appendix~\ref{appendix: examples}.

\subsection{Data Synthesis Framework}

Guided by the principles of high-quality synthetic multimodal data, \ie, broad scope, robust cross-modal alignment, and high fidelity, we introduce a data synthesis framework. This framework is designed to synthesize high-quality data that transforms an MLLM for downstream embedding tasks.

\subsubsection{Data Configuration}

To prepare for the data synthesis process, we configure the input data from three aspects:

\noindent \textbf{Task and Modality Combination}
We aim to synthesize data with a broad scope by generating beyond simple retrieval data of IT$\rightarrow$IT and T$\rightarrow$IT types.
Our data cover three key multimodal embedding tasks identified by previous work~\cite{MMEB}: classification, VQA, and retrieval.
After selecting a task for synthesis, we will sample a modality combination with respect to the specific task, such as choosing from seven possible combinations for the retrieval task type.
Note that we only synthesize data of modality types that are included in the MMEB benchmark~\cite{MMEB}, which can cover most scenarios.

\noindent \textbf{Image}
Despite the powerful multimodal capabilities of modern MLLMs (\eg, GPT-4o, Llama-3.2~\cite{llama3}, and Llava-1.6), most cannot generate images, and those that can often produce low-fidelity images~\cite{VISTA}.
Following previous works~\cite{MagicLens, megapairs}, we sample real images from the LAION-400M corpus~\cite{LAION}.
First, we will sample a query image from the corpus ($q_i \in \mathcal{I}$).
Then, for the modality types involving images on the document side (\eg, IT$\rightarrow$IT), we use a small embedding model, jina-clip-v2~\cite{jina-clip}, to retrieve a similar positive image $d^+_i$ and a hard negative image $d^-_i$ efficiently.

\noindent \textbf{Language}
Most existing models only focus on high-source languages like English, harming the multilingual ability of embedding models.
To synthesize multilingual data, we sample languages from the language list of XLM-R~\cite{XLM-R} during configuration.
In order to facilitate the common usage scenarios, we give high-source languages higher weights.
Note that the generated task instruction will always be in English for effective instruction tuning.

\subsubsection{One-pass Generation with MLLM}

With the data configuration ready, we introduce a deep thinking process that involves interpreting input images, generating data, and performing self-evaluation.
To ensure that the MLLM always takes the image context into account, we execute this entire process in a single pass.

\begin{table*}[h]
\centering
\small
\begin{tabular}{lccccccc}
\toprule
\multirow{2}{*}{{Models}}  & \multicolumn{4}{c}{{Per Meta-Task Score}}       &\multicolumn{3}{c}{{Average Scor}e} \\
\cmidrule(r){2-5}
\cmidrule(l){6-8}
 & Class. & VQA  & Retr. & Ground. &  IND  & OOD & Overall                     \\
\midrule
\multicolumn{7}{l}{\textit{Zero-shot Setting Models}}                   \\
CLIP~\cite{CLIP}                                                        & 42.8           & 9.1  & 53.0      & 51.8   & - & -  & 37.8                     \\
BLIP2~\cite{blip2}                                                       & 27.0           & 4.2  & 33.9      & 47.0  & - & -   & 25.2                     \\
SigLIP~\cite{SigLIP}                                                         & 40.3           & 8.4  & 31.6      & 59.5  & - & -   & 34.8                     \\
OpenCLIP~\cite{OpenCLIP}                                                & {47.8}           & 10.9 & 52.3      & 53.3   & - & -  & 39.7                     \\
E5-V~\cite{E5-V}                                           & 21.8           & 4.9  & 11.5      & 19.0    & - & - & 13.3    \\
MagicLens~\cite{MagicLens}                                                 & 38.8           & 8.3  & 35.4      & 26.0   & - & -   & 27.8                     \\
MMRet (w/ 26M synthetic data) & {47.2}           & {18.4} & \textbf{56.5}      & {62.2}      & - & -     & {44.0} \\
\rowcolor{gray!20}
{\ours{}} (w/ 560K synthetic data)             & \textbf{60.6}           & \textbf{55.7} & \underline{54.7}      & \textbf{72.4}      & - & -     & \textbf{58.6}  \\
\midrule
\multicolumn{7}{l}{\textit{Partially Supervised Finetuning Models}$^\dag$}     \\ 
UniIR~\cite{UniIR}                                   & 42.1           & {15.0} & {60.1}      & {62.2}  & -  & -  & {42.8}                     \\
MM-EMBED~\cite{mmembed}     & {48.1}           & {32.2} & {63.8}      & {57.8}      & - & -     & {50.0} \\
GME~\cite{GME}   & {56.9}              & {41.2} & {67.8}      & {53.4}      & - & -     & {55.8} \\
\midrule
\multicolumn{7}{l}{\textit{Supervised Finetuning Models}}                   \\
CLIP~\cite{CLIP}                     & 55.2           & 19.7 & 53.2      & 62.2      & 47.6     & 42.8     & 45.4        \\
OpenCLIP~\cite{OpenCLIP}                 & {56.0}           & 21.9 & 55.4      & 64.1      & 50.5     & 43.1     & 47.2        \\
VLM2Vec~\cite{MMEB}        & \underline{61.2}           & {49.9} & {67.4}      & \underline{86.1}      & {67.5}     & {57.1}     & {62.9}        \\
MMRet~\cite{megapairs} & {56.0}           & \underline{57.4} & \underline{69.9}      & {83.6}      & \underline{68.0}     & \underline{59.1}     & \underline{64.1} \\
\rowcolor{gray!20}
{\ours{}} (w/ synthetic data + labeled data)             & \textbf{67.6}           & \textbf{62.8} & \textbf{70.9}      & \textbf{89.7}      & \textbf{72.3}     & \textbf{66.7}     & \textbf{69.8}  \\

\bottomrule
\end{tabular}
\vspace{-0.1cm}
\caption{Results on MMEB benchmark, consisting of 36 tasks across four types: classification (Class.), VQA, retrieval (Retr.), and visual grounding (Ground.).
$^\dag$ UniIR, MM-EMBED, and GME are not strictly zero-shot models. 
UniIR and MM-EMBED are trained on the MBEIR dataset~\cite{UniIR}, which includes 10 retrieval datasets included in the MMEB. 
Similarly, GME is trained on the UMRB dataset~\cite{GME}, which shares 14 datasets with the MMEB.
For VLM2Vec, we use the LLaVA-based version with high-resolution images reported in its original paper.
The second-best performances are underlined and the best performances are in bold.
}
\label{tab:mmeb}
\end{table*}

\noindent \textbf{Multi-aspect Visual Interpretation}
To obtain a comprehensive understanding of the images, the MLLM $\pi_{\theta}$ first analyzes them from multiple perspectives:
(1) the general information,
(2) detailed description of the objects present,
(3) contextual background information, and
(4) potential connections between the image and the text that may be synthesized.
The deep understanding of the images enables $\pi_{\theta}$ to produce texts that are closely aligned with the visual content, thereby enhancing the cross-modal alignment.

\noindent \textbf{Synthesizing Data}
Using the images and their descriptions as input, we prompt $\pi_{\theta}$ to synthesize texts $(t, q_t, d^+_t, d^-_t)$.
Specifically, the text instruction $t$ is expected to connect $q_i$ with $d^+_i$.\footnote{Because of limited space, full prompts are omitted in this section. The complete prompts can be found in Appendix~\ref{appendix: prompt}.}
The query and document texts should be relevant to their respective images.
Note that the input and output formats for the synthetic data may vary depending on the combination of modalities.
For example, for I$\rightarrow$IT and T$\rightarrow$IT types, there can be no query text and image, respectively.

\noindent \textbf{Self-evaluation}
In order to further enhance the quality of the synthetic data, $\pi_{\theta}$ evaluates the data it synthesizes from:
(1) the relevance of the texts to their corresponding images,
(2) the plausibility of hard negatives,
(3) the clarity of $t$, and
(4) the diversity (creativity) of the synthesized data.
Following this evaluation, $\pi_{\theta}$ provides suggestions for potential improvements.
Finally, a revised version of each data sample is produced and utilized for the subsequent contrastive training phase.

\subsection{Finetuning Embedding Model \ours{}}

Following previous works of instruction-tuned text embedding models~\cite{bge, Llama2Vec} and multimodal embedding models~\cite{MMEB}, we apply an instruction template on each query: $ \text{[IMAGE]}~\{t\}~\textbackslash n~\{q_t\}~\{q_i\}$, where ``$\text{[IMAGE]}$'' is the image token that varies from different MLLMs.
We then append an ``$\text{[EOS]}$'' token to each query and document.
The representation of each input in an MLLM is derived from the output of the ``$\text{[EOS]}$'' token from the final layer.

We utilize the InfoNCE loss~\cite{infonce} to perform the standard contrastive learning objective on our synthetic data $\mathcal{D}$:
\begin{eqnarray}
\mathcal{L} = -\log\frac{\phi(\mathbf{q},\mathbf{d}^+)}{\phi(\mathbf{q},\mathbf{d}^+) + \sum_{{d}^-\in\mathcal{N}}{\phi(\mathbf{q},\mathbf{d}^-)}}, 
\label{equation:cl}
\end{eqnarray}
where $\mathbf{q}$ is the encoded multimodal query, $\mathbf{d}$ represents the encoded document, and $\mathcal{N}$ denotes the set of negative documents.
The function $\phi(\cdot) = \exp(\text{cos}(\cdot) / \tau)$, where ${\rm cos}(\cdot)$ denotes cosine similarity, and $\tau$ is a temperature hyperparameter.

%% file: sections/4_experiments.tex
\section{Experiments}

\subsection{Experimental Setup}
\label{subsec:setup}

We synthesize a total of 560K multimodal embedding data samples.
The MLLM utilized for data synthesis is \textit{GPT-4o-2024-08-06}.
The backbone model for \ours{} is Llama-3.2-11B-Vision\footnote{\url{https://huggingface.co/meta-llama/Llama-3.2-11B-Vision}}.
For finetuning \ours{}, we employed LoRA~\cite{lora} with a rank of 8.
We evaluate the general embedding performance in terms of Precision@1 on the MMEB benchmark~\cite{MMEB}.
This benchmark comprises 36 multimodal embedding tasks across four categories: classification (10), VQA (10), retrieval (12), and visual grounding (4).
Our synthetic dataset is distributed among classification, VQA, and retrieval tasks in a 1:1:2 ratio.
We synthesize more retrieval data since this type contains more kinds of modality combinations.
We do not synthesize visual grounding data since they are relatively simpler for MLLM based on the MMEB results.
To evaluate multilingual multimodal capabilities, we conducted tests using the XTD benchmark~\cite{XTD}.
Following MURAL~\cite{mural}, we conduct experiments on seven languages of XTD and report Recall@10 results.
Additional details regarding the synthetic data, prompts, and implementation can be found in Appendix~\ref{appendix: syndata}, \ref{appendix: implementation}, and~\ref{appendix: prompt}, respectively.

\begin{figure}[!tbp]
\centering
\includegraphics[width=0.45\textwidth]{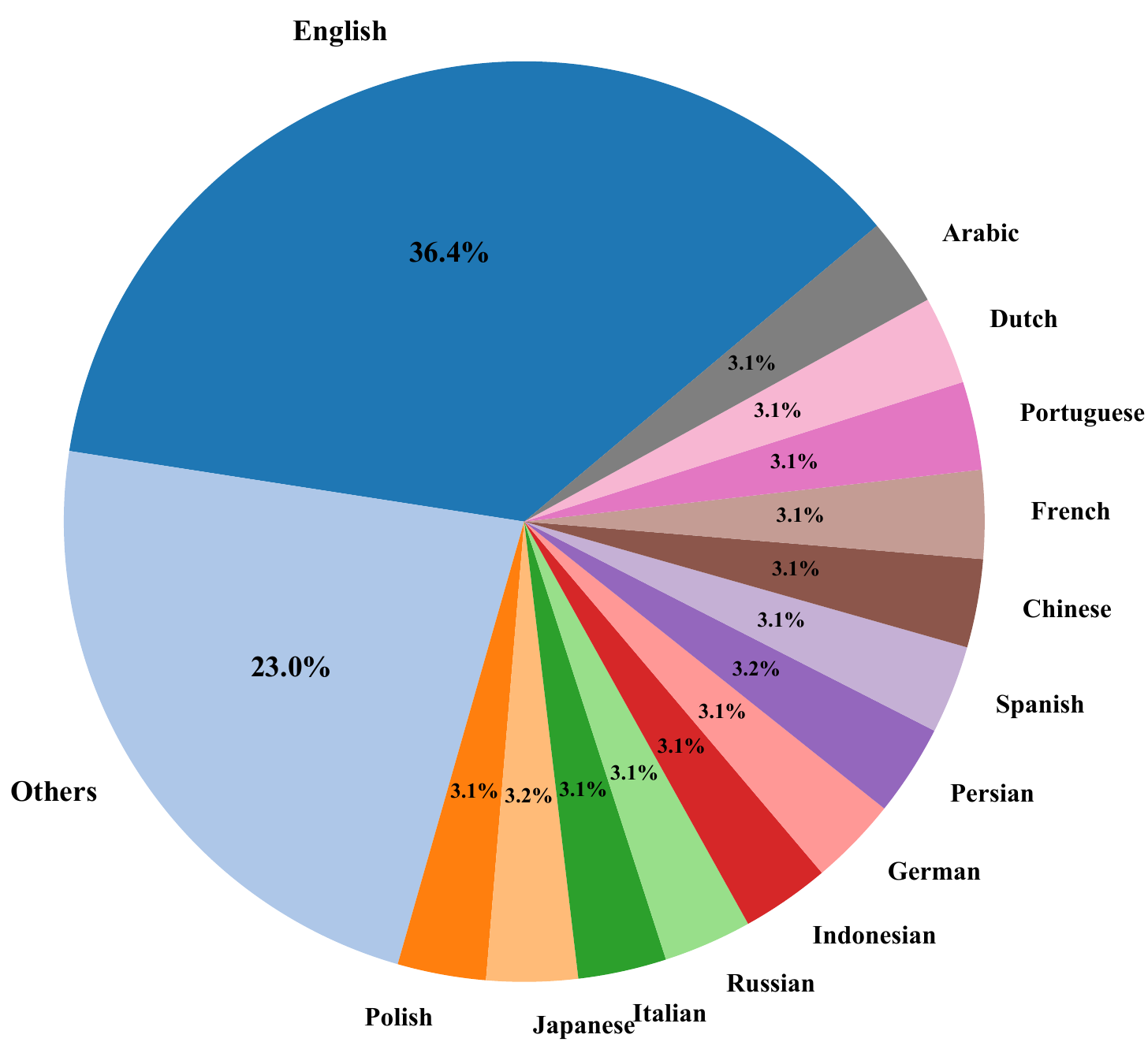}
\vspace{-2pt}
\caption{Distribution of languages in the synthetic data.}
\label{fig:language_dist} 
\end{figure}

\subsection{Results on MMEB}

The overall results on the MMEB benchmark are presented in Table~\ref{tab:mmeb}.
\ours{} achieves the best performance on both zero-shot setting (with synthetic data only) and supervised setting (with IND training datasets of MMEB).
This demonstrates the quality of our synthetic data and the effectiveness of our multimodal embedding model.
Furthermore, we can make the following observations:
(1) \ours{} generalizes well on all four kinds of tasks.
This demonstrates the broad scope of our synthetic multimodal embedding data in terms of task types.
(2) With only 560K synthetic data, \ours{} manages to perform better than MMRet which uses 26M data.
This proves the quality of our synthetic data again.
(3) Intriguingly, \ours{} underperforms MMRet on retrieval tasks in a zero-shot setting.
This is because MMRet is trained on 26M pure retrieval data, which makes it perform well on retrieval tasks, but generalizes poorly on other task types.

\subsection{Multilingual Performance on XTD}

We synthesize a multilingual multimodal dataset that consists of 93 languages, in order to train our embedding model \ours{} to generalize across more languages.
The language distribution of our dataset is presented in Figure~\ref{fig:language_dist}.
Notably, the dataset primarily consists of English data samples, facilitating common usage scenarios.
For the 75 low-resource languages, we evenly synthesize data samples to obtain a balanced multilingual dataset that supports comprehensive cross-linguistic generalization.

To evaluate the multilingual capability of \ours{}, we conduct experiments across seven languages on a text-to-image retrieval benchmark XTD.
As presented in Table~\ref{tab:xtd}, \ours{} outperforms other models in terms of overall performances on all languages, demonstrating its superior multilingual multimodal embedding capability.
The following observations can be made:
(1) The multilingual performance of multimodal embedding models is largely dependent on their foundational models.
For example, jina-clip-v2 and M-CLIP outperform VLM2Vec-LLaVA, despite VLM2Vec's strong performance on MMEB.
GME exhibits robust performance on XTD, which can be attributed to the powerful multilingual MLLM, Qwen2-VL~\cite{Qwen2-VL}.
(2) The performance of \ours{} declines when labeled data is omitted, indicating that general multimodal capabilities remain essential for multilingual retrieval tasks.
(3) In a zero-shot setting, \ours{} trained on multilingual synthetic data (\ours{} w/ synthetic data only) outperforms \ours{} with the same amount of English synthetic data (\ours{} w/ english synthetic data).
This suggests that the extensive language coverage provided by our synthetic data enhances the multilingual capabilities of embedding models.

\begin{table}[!tbp]
\centering
\scriptsize
\setlength{\tabcolsep}{2pt}
\begin{tabular}{l|cccccccc}
\toprule
Model          & it   & es   & ru   & zh   & pl   & tr   & ko   & Avg. \\
\midrule
ALIGN~\cite{ALIGN}           & 87.9 & 88.8 & 82.3 & 86.5 & 79.8 & 73.5 & 76.6 & 82.2 \\
MURAL~\cite{mural}           & 91.8 & 92.9 & 87.2 & 89.7 & 91.0 & 89.5 & 88.1 & 90.0 \\
VLM2Vec~\cite{MMEB} & 83.7 & 87.1 & 86.7 & 92.8 & 76.1 & 37.2 & 63.9 & 75.4 \\
jina~\cite{jina-clip}    & 93.6 & 94.1 & 89.8 & 91.8 & 94.3 & 92.7 & 90.1 & 92.3 \\
M-CLIP~\cite{mclip}  & 93.1 & 93.6 & 90.0 & 94.0 & 94.3 & 93.1 & 89.0 & 92.4 \\
GME~\cite{GME}             & 95.1 & 96.4 & 92.3 & 96.4 & 94.9 & 89.8 & 93.6 & 94.1 \\
\midrule
\ours{} (full) & 96.1 & 96.2 & 93.3 & 96.3 & 95.4 & 93.6 & 96.0 & \textbf{95.3} \\
\quad w/ synthetic data only & 90.9 & 89.6 & 86.3 & 90.2 & 90.3 & 87.2 & 86.7 & 88.7 \\
\quad w/ english synthetic data & 86.3 & 86.3 & 84.2 & 88.8 & 84.9 & 81.0 & 84.4 & 85.1 \\
\bottomrule
\end{tabular}
\caption{Results on XTD benchmark, a text-to-image retrieval task covering seven languages.
}
\label{tab:xtd}
\end{table}

\subsection{Application to Other Base MLLM}

We train \ours{} based on the powerful MLLM LLaMA-3.2-Vision, which is instruction-tuned and effective in interpreting multimodal inputs.
Notably, our synthetic data and training paradigm can effectively transform other foundation MLLMs into embedding models.
We use both our synthetic data and labeled data to train LLaVA-1.6\footnote{\url{https://huggingface.co/llava-hf/llava-v1.6-mistral-7b-hf}} and Phi-3.5-V\footnote{\url{https://huggingface.co/microsoft/Phi-3.5-vision-instruct}}.
The performances of \ours{} with different foundation MLLMs are presented in Table~\ref{tab:base_model}.
The results show that models trained using our method consistently outperform baseline models built on the same foundational MLLMs.
This indicates that our synthetic data can effectively enhance the capability of MLLMs to embed multimodal inputs.

\begin{table}[!t]
    \centering
    \small
    \begin{tabular}{p{0.3\textwidth}cc}
    \toprule
        Base MLLM & Avg. on MMEB  \\
        \midrule
        Phi-3.5-V~\cite{Phi3} &   61.0 \\
        LLaVA-1.6~\cite{Llava} &  65.8  \\
        LLaMA-3-Vision~\cite{llama3} (Ours)  & \textbf{69.8}  \\
    \midrule
    \multicolumn{2}{l}{\textit{Baselines (For Reference)}} \\ 
    \midrule
        VLM2Vec (Phi-3.5-V) &   60.1 \\
        VLM2Vec (LLaVA-1.6) &   62.9 \\
        MMRet (LLaVA-1.6) &   64.1 \\
        VLM2Vec (LLaMA-3.2) &   64.8 \\
    \bottomrule
    \end{tabular}
    \caption{Performances of \ours{} with different MLLMs.}
    \label{tab:base_model}
\end{table} 

\subsection{Discussions of Data Synthesis Process}

In this section, we will further investigate the data synthesis process via zero-shot experiments.

\subsubsection{Ablation Studies}

To evaluate each component of our data synthesis framework, we conduct ablation studies of \ours{}:

\noindent \textbf{Deep Thinking Process}
To synthesize high-quality data, we introduce a deep thinking process to boost data synthesis.
As presented in Table~\ref{tab:ablation}, the performance of \ours{} declines when the Visual Interpretation and Self-evaluation components are excluded.
For example, \ours{} performs worse when utilizing the original data compared to revised data.
This indicates that the self-evaluation mechanism can enhance data fidelity, facilitating the training of a more robust embedding model.

\begin{table}[!t]
    \centering
    \small
    \renewcommand{\arraystretch}{1.3}
    \begin{tabular}{lc}    
    \toprule
         Model & Avg. on MMEB  \\
        \midrule
        \ours{} (280K synthetic data only) & \textbf{57.4}  \\
        \midrule
        \quad w/o. Visual Interpertation &  57.2  \\
        \quad w/o. Self-evaluation &  56.0  \\
        \hline
        \quad w/o. Classification Data &  52.5  \\
        \quad w/o. VQA Data &  55.1  \\
        \quad w/o. Retrieval Data &  56.5  \\
        \hline
        \quad w/ IT2I only (MagicLens~\&~MegaPairs) & 30.1   \\
        \quad w/  IT2IT~\&~T2IT only (GME) &  28.6  \\
        \hline
        \quad w/o. Hard Negative &  56.2  \\
        \hline
        \quad w/ English Data only (280K) & 57.6 \\
        \quad w/o. English Data (280K) & 56.9 \\
    \bottomrule
    \end{tabular}
    \caption{Performances of ablated models on MMEB.
For efficient test, we conduct zero-shot experiments on 280K synthetic data, which has the same tasks, modality types and languages as the full synthetic data.}
    \label{tab:ablation}
\end{table} 

\noindent \textbf{Embedding Task Types}
In order to expand the scope of data, we synthesize data across three task types: classification, VQA, and retrieval.
The performance of \ours{} decreases after each type of multimodal embedding data is omitted, demonstrating that our diverse synthetic data can facilitate model generalization.
Intriguingly, the performance drops the least after removing the retrieval data, which is inconsistent with previous research~\cite{MMEB}.
One possible explanation is that our backbone, Llama-3.2 Vision, inherently exhibits more robust retrieval capabilities than Phi-3.5-V. 

\noindent \textbf{Modality Combinations}
Most prior works focus on one or two modality types, such as ``IT2I'' (\eg, MagicLens~\cite{MagicLens} and MegaPairs~\cite{megapairs}) or ``IT2IT~\&~T2IT'' (\eg, GME~\cite{GME}).
We propose to synthesize data across various modality combinations to enhance the diversity of our synthetic dataset, \ie, the scope of our synthetic multimodal data.
To evaluate the impact of these additional modality combinations, we train \ours{} with the same amount of datasets that contain types ``IT2I'' or ``IT2IT~\&~T2IT'' only.
The performance of \ours{} significantly decreased when limited to these combinations from previous works, which indicates that the additional modalities enable our embedding model to generalize more effectively across different combinations and task types.

\noindent \textbf{Hard Negative}
Each sample in our synthetic dataset incorporates a hard negative document to help \ours{} learn subtle differences.
After excluding the hard negatives, the model's performance drops significantly, which demonstrates the importance of this technique for contrastive learning.

\noindent \textbf{Language}
To investigate the impact of linguistic diversity on model performance on English benchmarks, we conducted experiments using synthetic data in two configurations: English-only and non-English languages only. 
Our model, \ours{}, demonstrated a slight performance advantage with English-only synthetic data, although the difference was minimal. 
Nonetheless, \ours{} achieved satisfactory results with 280K data samples from languages other than English. 
This suggests that our multilingual dataset enhances the embedding model's ability to generalize effectively in both multilingual and English-only contexts.

\subsubsection{Scaling Effect}

The scaling effect is an important aspect of synthetic data generation for multimodal embedding models~\cite{GME, megapairs}. 
It explores how the performance of the model varies with the size of synthetic datasets. 
Besides, the data synthesis and training processes demand significant computational resources and time.
Therefore, studying the scaling effect allows us to identify the point of diminishing returns, ensuring that resources are utilized efficiently without overproducing redundant data.

\begin{figure}[t]
\centering
\includegraphics[width=0.45\textwidth]{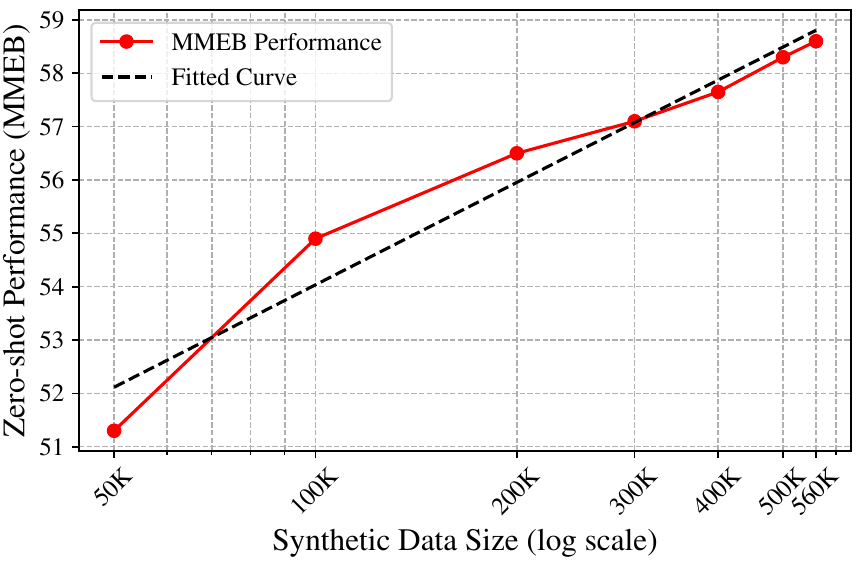}
\vspace{-2pt}
\caption{The impact of synthetic data size on multimodal embedding performance on MMEB.}
\label{fig:scaling_law} 
\end{figure}

In this section, we further investigate the performance of \ours{} using synthetic datasets of varying sizes.
Specifically, we conduct zero-shot experiments on MMEB to analyze the scaling effect.
As illustrated in Figure~\ref{fig:scaling_law}, \ours{} consistently achieves better performance with increased training data, demonstrating the high quality of our synthetic data again.
This paradigm also indicates a linear-log relationship between the model performance and data size, consistent with previous works of text embedding~\cite{speed} and dense retrieval~\cite{scaling_ds}.
This finding facilitates the balancing of the cost and the multimodal embedding model performance for future works.

\subsection{Hyperparameter Analysis}

In order to analyze the training process of our multimodal embedding model, we perform experiments with \ours{} using various training settings.
For efficiency, we report zero-shot results for \ours{} trained with 280K synthetic data.
Note that we tune these hyperparameters on evaluation datasets comprising 1K samples from each training set.
However, for consistency with previous experiments, we present results on the MMEB test sets.

\begin{figure}[t]
\centering
\includegraphics[width=0.47\textwidth]{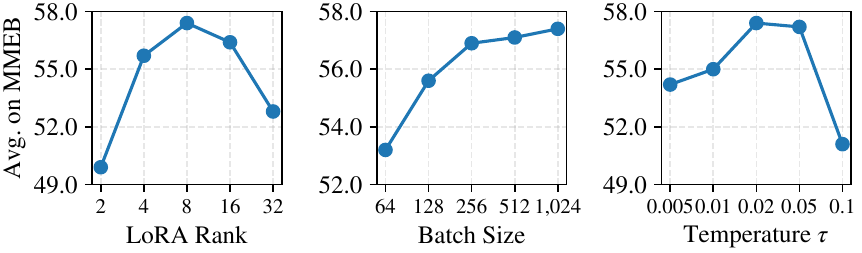}
\vspace{-2pt}
\caption{The zero-shot performances of \ours{} with different training settings on MMEB (280K synthetic data for efficient test).}
\label{fig:param} 
\end{figure}

\noindent \textbf{LoRA Rank}
denotes the rank of the additional low-rank matrices in LoRA. 
This parameter influences the number of parameters added into the original model, balancing the model's capacity and computational efficiency.
As shown in the left part of Figure~\ref{fig:param}, the performance of \ours{} initially improves then drops.
This demonstrates a trade-off: a lower rank reduces memory and computation but may lead to underfitting if r is too small, whereas a higher rank risks harming the pre-trained multimodal reasoning capabilities of MLLM.

\noindent \textbf{Training Batch Size}
In contrastive learning, batch size plays a critical role because it directly affects the number of negative samples available for training.
As presented in the middle part of Figure~\ref{fig:param}, the performance of \ours{} consistently increases with larger batch size.
However, large batches demand significantly more GPU memory, \ie, more computational resources.

\noindent \textbf{Temperature }
The temperature parameter $\tau$ in the InfoNCE loss (Equation~\ref{equation:cl}) influences the separation between positive and negative samples in the embedding space.
We can observe that \ours{}'s performance first improves then declines with larger temperature.
This pattern suggests a trade-off:
a low $\tau$ forces the model to strongly penalize near-positive negatives which can lead to overfitting, while a high $\tau$ leads to a more uniform distribution of embeddings which may hinder the effective separation of positive and negative samples.

%% file: sections/5_conclusion.tex
\section{Conclusion}
In this work, we synthesize high-quality multimodal multilingual data to train the model \ours{}.
We first define high-quality multimodal synthetic data based on three criteria: broad scope, robust cross-modal alignment, and high fidelity.
Then, we develop a data synthesis framework guided by these principles.
Finally, we train a multimodal multilingual embedding model using the high-quality synthetic data.
\ours{} achieves SOTA performances on both the general benchmark MMEB and the multilingual benchmark XTD.


\section*{Limitations}

Our work has several limitations that we intend to resolve in future research:

\begin{enumerate}
    \item Our model currently relies on the proprietary MLLM GPT-4o for synthesizing multimodal data. Future work should explore aligning smaller MLLMs with the knowledge from GPT-like models to achieve more efficient data synthesis.
    \item \ours{} focus on text and image modalities. Future models should aim to extend coverage to additional modalities, such as audio and video.
    \item Due to the cost limitation and the observed scaling effect, we limited the amount of data produced for model training. Future research may consider increasing data size while preserving diversity to optimize model performance.
\end{enumerate}

%% file: sections/appendix.tex
\clearpage

\section*{Appendix}

\section{Details about Synthetic Data}
\label{appendix: syndata}

\begin{table}[h]
\small
\centering
\begin{tabular}{l|c|c}
\toprule
Task                            & Modality Combination                    & \# samples       \\ \midrule
\multirow{2}{*}{Classification} & image-to-text                &          126,177             \\
                                & (image,text)-to-text         &     13,823                  \\
\midrule
\multirow{7}{*}{Retrieval}      & image-to-text                &   98,040                    \\
                                & (image,text)-to-text         &      41,960                 \\
                                & (image,text)-to-image        &     56,185                  \\
                                & image-to-image               &     27,988                  \\
                                & (image,text)-to-(image,text) &   27,656                    \\
                                & text-to-image                &    14,090                   \\
                                & text-to-(image,text)         &  14,081    \\
\midrule
VQA                             & (image,text)-to-text         & 140,000 \\
\bottomrule
\end{tabular}
\caption{Statistics of the multimodal synthetic data used for training \ours{}.} 
\label{tab:syndata_statistics}
\end{table}

In this study, we introduce a synthetic multimodal multilingual embedding dataset designed to facilitate model learning. 
This section delves into the details of our synthetic dataset. 
The dataset is comprised of three distinct tasks and seven modality combinations, totaling 560K data samples. 
Table~\ref{tab:syndata_statistics} provides a detailed statistical overview of our synthetic data, categorized by tasks and modalities.

\section{Implementation Details}
\label{appendix: implementation}

\subsection{Data Synthesis}

For the data synthesis process, we employ the MLLM \textit{GPT-4o-2024-08-06} model to generate data samples. 
Both the temperature and top-p parameters are set to 1.0 to ensure diverse and coherent outputs. 
Our image corpus is sourced from LAION-400m~\cite{LAION}, from which we exclude images that are either corrupted or have inaccessible URLs. 
Each synthetic data sample incorporates one image sampled from this corpus as the query image. 
For modality combinations that include images on the document side, we utilize the jina-clip-v2\footnote{\url{https://huggingface.co/jinaai/jina-clip-v2}} model to retrieve a similar image, along with a hard negative image, to serve as additional inputs.

\subsection{Finetuning Embedding Model}

We train \ours{} using the open-source MLLM, Llama-3.2-11B-Vision\footnote{\url{https://huggingface.co/meta-llama/Llama-3.2-11B-Vision}}. 
The training is conducted on 64 NVIDIA A100 GPUs, each equipped with 40GB of memory. 
To optimize GPU memory usage, we employ gradient checkpointing and set the gradient accumulation steps to 4.
The model is trained with a learning rate of 2e-5 for one epoch, utilizing both synthetic and labeled data. 
LoRA~\cite{lora} is applied to the MLLM with a rank of 8. 
Each training sample incorporates one hard negative document. 
Hard negatives are mined for each subset of MMEB using VLM2Vec-LoRA\footnote{\url{https://huggingface.co/TIGER-Lab/VLM2Vec-LoRA}}, with the 70th position in the ranking list selected as the hard negative sample.

More implementation details can be found in \url{https://github.com/haon-chen/mmE5}.

\section{Prompts}
\label{appendix: prompt}

We use different prompts of data synthesis for different tasks.
For retrieval task, we design two prompts for modality combinations that involve images on the document side or not.
Let us take the prompt of generating classification data for an example to illustrate the prompt design.

First, we sample a modality combination from \{image-to-text, (image,text)-to-text\}.
If the query side does not include texts, the ``input\_text'' of the classification data sample will be an empty string.
Similarly, for modalities of retrieval task that do not include document texts, the ``positive\_document'' and ``hard\_negative\_document'' will be empty.
Following previous works of synthesizing text embedding data~\cite{E5mistral, speed}, we will randomly select a clarity and difficulty setting to enhance diversity.

Then, for the multi-aspect visual description process, we ask the MLLM to explicitly \textcolor{red}{include four perspectives of description.}
Besides, for the data synthesis process, we also ask the MLLM to \textcolor{red}{follow some specific guidelines.}  
Furthermore, the MLLM will \textcolor{red}{evaluate the initially generated data from several aspects} and provide ``possible\_improvements''.
Finally, the revised version of data will be used as the output data sample.
Note that there are no task instructions generated for the VQA task, since they are all fixed as ``Represent the given image with the following question:''.

\begin{figure*}[p]
\centering
\begin{prompt}[title={Prompt: Synthesizing Classification Data}, label=prompt:cla]

Your mission is to first produce a detailed visual description of the image (within 300 words), identifying all potential aspects for generating high-quality data for a \textcolor{blue}{\{image-to-text, (image,text)-to-text\}} classification task. \\
    
Based on the description, brainstorm a potentially useful task. \\

Here are a few examples for your reference:
\textcolor{blue}{\{example tasks\}} \\

Then, you should write one multi-modal classification example for this task in JSON format. The JSON object must contain the following keys: \\

- "description": a string, your detailed visual description, listing all required elements. \\
- "task\_instruction": a string, describing the classification task. \\
- "input\_text": \textcolor{blue}{\{"an empty string", "a string the input text specified by the classification task"\}}. \\
- "label": a string, the correct label of the image and input\_text (if not empty) based on the task instruction. \\
- "misleading\_label": a string, an incorrect label that is related to the task. \\
- "evaluation": a string, a brief summary of the evaluation of data quality. \\
- "possible\_improvements": a string, suggestions for improving the data based on the guidelines. \\
- "revised\_task\_instruction": the revised task instruction. \\
- "revised\_input\_text": the revised input text, \textcolor{blue}{\{"an empty string", "a string the input text specified by the classification task"\}}. \\
- "revised\_label": the revised label. \\
- "revised\_misleading\_label": the revised misleading label. \\

\textcolor{red}{For the description, please include the following elements:}\\
- General Description: Provide an overall summary of the image, including the primary objects, scene, and notable features.\\
- Object-Level Details: Identify the individual objects in the image, their attributes (e.g., color, size, position), and their relationships to one another.\\
- Contextual Features: Describe the scene or environment, including background details, lighting, and any actions taking place.\\
- Task-specific Brainstorming: Analyze explore how this image could relate to text (e.g., captions, contextual descriptions).\\

\textcolor{red}{Please adhere to the following guidelines:}\\
- Task should be suitable for the given image.\\
- Avoid generate task similar to classification of sentiment / subject / study field / genre / main topic / spam / urgency / language.\\
- The "input\_text" should be \textcolor{blue}{\{"less than 10", "at least 10", "at least 50", "at least 100", "at least 200"\}} words and diverse in expression (if not empty).\\
- The "misleading\_label" must be a valid label for the given task, but not as appropriate as the "label" for the image.\\
- The text of "task\_instruction" should be in English and others fields should be in \textcolor{blue}{\{language\}}.\\
- Avoid including the values of the "label" and "misleading\_label" fields in the "input\_text" (if not empty), that would make the task too easy.\\
- The "input\_text" (if not empty) is \textcolor{blue}{\{"clear", "understandable with some effort", "ambiguous"\}} and requires \textcolor{blue}{\{"high school", "college", "PhD"\}} level education to comprehend.\\
- \textcolor{red}{When generating the data, please evaluate the following aspects:}\\
  1. Relevance: Are the generated input texts and labels (if not empty) tightly connected to their corresponding image and task objectives? Does the task instruction effectively link the query image with the positive label?\\
  2. Plausibility: Are misleading labels sufficiently relevant to the image or labels while remaining definitively incorrect? Could they mislead the model?\\
  3. Clarity: Is the generated task clear and unambiguous, providing sufficient instruction to connect the query image with the label, without being overly specific or abstract?\\
  4. Diversity: Does the generated data introduce variation in task instructions, texts (if not empty), and labels to avoid repetitive patterns in the dataset?\\
- Provide a detailed evaluation of the data based on the above criteria. For each criterion, explain specific flaws or strengths.\\
- Suggest specific revisions to address any identified weaknesses, ensuring the revised data better aligns with the guidelines and task objectives.\\
- Avoid revisions that overly simplify the task instruction, text (if not empty), or labels, as this may reduce their utility for training.\\
- Ensure that revised data maintains consistency with the corresponding image content and classification task requirements.\\

Your output must always be a JSON object only. Do not explain yourself or output anything else. Be creative!
    
\end{prompt}
\end{figure*}

\begin{figure*}[t]
\centering
\begin{prompt}[title={Prompt: Synthesizing VQA Data}, label=prompt:vqa]

Your mission is to first produce detailed visual descriptions of the image (within 300 words), identifying all potential aspects for generating high-quality data for a visual QA task. \\
    
Based on the description, write one visual QA example based on the given image in JSON format. The JSON object must contain the following keys: \\

- "description": a string, your detailed visual description, listing all required elements.\\
- "question": a string, specifying the question based on the image content.\\
- "positive\_answer": a string, the correct answer for the question based on the image content.\\
- "hard\_negative\_answer": a string, an incorrect answer that appears plausible but is ultimately wrong.\\
- "evaluation": a string, a brief summary of the evaluation of data quality.\\
- "possible\_improvements": a string, suggestions for improving the data based on the guidelines.\\
- "revised\_question": the revised question.\\
- "revised\_positive\_answer": the revised positive answer.\\
- "revised\_hard\_negative\_answer": the revised hard negative answer.\\

\textcolor{red}{For the description, please include the following elements:}\\
- General Description: Provide an overall summary of the image, including the primary objects, scene, and notable features.\\
- Object-Level Details: Identify the individual objects in the image, their attributes (e.g., color, size, position), and their relationships to one another.\\
- Contextual Features: Describe the scene or environment, including background details, lighting, and any actions taking place.\\
- Task-specific Brainstorming: Analyze explore how this image could relate to text (e.g., captions, contextual descriptions).\\

\textcolor{red}{Please adhere to the following guidelines:}\\
- The "question" should be \textcolor{blue}{\{"less than 10", "at least 10", "at least 50", "at least 100", "at least 200"\}} words and diverse in expression.\\
- The "hard\_negative\_answer" must be plausible but less appropriate than the "positive\_answer".\\
- The values for all fields should be in \textcolor{blue}{\{language\}}.\\
- Avoid including explicit hints in the question that make the answer too obvious.\\
- The "question" (if not empty) is \textcolor{blue}{\{"clear", "understandable with some effort", "ambiguous"\}} and requires \textcolor{blue}{\{"high school", "college", "PhD"\}} level education to comprehend.\\
- \textcolor{red}{When generating the data, please evaluate the following aspects:}\\
  1. Relevance: Are the generated question and answers tightly linked to the image content and consistent with the task requirements?\\
  2. Plausibility: Does the "hard\_negative\_answer" closely resemble the "positive\_answer" while remaining definitively incorrect? Could it mislead the model?\\
  3. Diversity: Does the generated data introduce variation in questions, and answers to avoid repetitive patterns in the dataset?\\
- Provide a detailed evaluation of the data based on the above criteria. For each criterion, explain specific flaws or strengths.\\
- Suggest specific revisions to address any identified weaknesses, ensuring the revised data better aligns with the guidelines and task objectives.\\
- Avoid revisions that overly simplify or trivialize the "question".\\
- Ensure revised data maintain consistency with the image content and task-specific requirements.\\

Your output must always be a JSON object only. Do not explain yourself or output anything else. Be creative!
    
\end{prompt}
\end{figure*}

\begin{figure*}[t]
\centering
\begin{prompt}[title={Prompt: Synthesizing Retrieval Data (Only Query Image)}, label=prompt:ret_one_image]

Your mission is to first produce a detailed visual description of the image (within 300 words), identifying all potential aspects for generating high-quality data for a \textcolor{blue}{\{image-to-text, (image,text)-to-text\}} retrieval task.\\
    
Based on the description, brainstorm a potentially useful task. \\

Here are a few examples for your reference:
\textcolor{blue}{\{example tasks\}} \\

Then, you should write one retrieval example for this task in JSON format. The JSON object must contain the following keys: \\

- "description": a string, your detailed visual description, listing all required elements. \\
- "task\_instruction": a string, describing the retrieval task. \\
- "query": \textcolor{blue}{\{"an empty string", "a random user search query specified by the retrieval task and the query image."\}} \\
- "positive\_document": a string, the relevant document for the query image content. \\
- "hard\_negative\_document": a string, a hard negative document that only appears relevant to the query image content. \\
- "evaluation": a string, a brief summary of the evaluation of data quality. \\
- "possible\_improvements": a string, suggestions for improving the data based on the guidelines. \\
- "revised\_task\_instruction": the revised task instruction. \\
- "revised\_query": the revised query, \textcolor{blue}{\{"an empty string", "a random user search query specified by the retrieval task and the query image."\}}. \\
- "revised\_positive\_document": the revised positive document, a string, the relevant document for the query image content. \\
- "revised\_hard\_negative\_document": the revised hard negative document, a string, a hard negative document that only appears relevant to the query image content. \\

\textcolor{red}{For the description, please include the following elements:}\\
- General Description: Provide an overall summary of the image, including the primary objects, scene, and notable features.\\
- Object-Level Details: Identify the individual objects in the image, their attributes (e.g., color, size, position), and their relationships to one another.\\
- Contextual Features: Describe the scene or environment, including background details, lighting, and any actions taking place.\\
- Task-specific Brainstorming: Analyze explore how this image could relate to text (e.g., captions, contextual descriptions).\\

\textcolor{red}{Please adhere to the following guidelines:}

- The task should involve both query and documents (positive and hard negative, if not empty). It must directly indicate the relation without being overly detailed or abstract.\\
- The query (if not empty) should be \textcolor{blue}{\{"extremely long-tail", "long-tail", "common"\}}, \textcolor{blue}{\{"less than 5 words", "5 to 15 words", "at least 10 words"\}}, \textcolor{blue}{\{"clear", "understandable with some effort", "ambiguous"\}}, and diverse in topic.\\
- All documents (if not empty) must be created independent of the query. Avoid copying the query verbatim. It’s acceptable if some parts of the "positive\_document" are not topically related to the query.\\
- All documents (if not empty) should be at least \textcolor{blue}{\{"10", "30", "200", "300"\}} words long.\\
- The "hard\_negative\_document" (if not empty) contains some useful information, but it should be less useful or comprehensive compared to the "positive\_document".\\
- The text of "task\_instruction" should be in English and others fields should be in \textcolor{blue}{\{language\}}.\\
- Do not provide any explanation in any document (if not empty) on why it is relevant or not relevant to the query.\\
- Do not use the word "query" or "document" in the generated content.\\
- Both the query and documents (if not empty) require \textcolor{blue}{\{"high school", "college", "PhD"\}} level education to understand.\\
- \textcolor{red}{When generating the data, please evaluate the following aspects:}\\
  1. Relevance: Are the generated query and documents (if not empty) tightly connected to their corresponding image and task objectives? Does the task instruction effectively link the query image with the positive text?\\
  2. Plausibility: Are hard negatives sufficiently similar to the query or positive examples while remaining definitively incorrect? Could they mislead the model?\\
  3. Clarity: Is the generated task clear and unambiguous, providing sufficient instruction to connect the query image with the positive document, without being overly specific or abstract?\\
  4. Diversity: Does the generated data introduce variation in task instructions, queries, and documents to avoid repetitive patterns in the dataset?\\
- Provide a detailed evaluation of the data based on the above criteria. For each criterion, explain specific flaws or strengths.\\
- Suggest specific revisions to address any identified weaknesses, ensuring the revised data better aligns with the guidelines and task objectives.\\
- Avoid revisions that overly simplify the task instruction, query, or documents, as this may reduce their utility for training.\\
- Ensure that revised data maintains consistency with the corresponding image content and retrieval task requirements.\\

Your output must always be a JSON object only. Do not explain yourself or output anything else. Be creative!
    
\end{prompt}
\end{figure*}

\begin{figure*}[t]
\centering
\begin{prompt}[title={Prompt: Synthesizing Retrieval Data (With Document Images)}, label=prompt:ret_three_image]
Your mission is to first produce detailed visual descriptions of the images (within 600 words), identifying all potential aspects for generating high-quality data for a \textcolor{blue}{\{(image,text)-to-image, image-to-image, (image,text)-to-(image,text), text-to-image, text-to-(image,text)\}} retrieval task that involves both query and document images.\\
    
Based on the description, brainstorm a potentially useful task. \\

Here are a few examples for your reference:
\textcolor{blue}{\{example tasks\}} \\

Then, you should write one retrieval example for this task in JSON format. The JSON object must contain the following keys: \\

- "description": a string, your detailed visual description, listing all required elements. \\
- "task\_instruction": a string, describing the retrieval task. \\
- "query": \textcolor{blue}{\{"an empty string", "a random user search query specified by the retrieval task and the query image."\}} \\
- "positive\_document": \textcolor{blue}{\{"an empty string", "a string, the relevant document for the query based on the query text and image content"\}} \\
- "hard\_negative\_document": \textcolor{blue}{\{"an empty string", "a string, a hard negative document that only appears relevant to the query and the query image content."\}} \\
- "evaluation": a string, a brief summary of the evaluation of data quality. \\
- "possible\_improvements": a string, suggestions for improving the data based on the guidelines. \\
- "revised\_task\_instruction": the revised task instruction. \\
- "revised\_query": the revised query, \textcolor{blue}{\{"an empty string", "a random user search query specified by the retrieval task and the query image."\}}. \\
- "revised\_positive\_document": the revised positive document, a string, \textcolor{blue}{\{"an empty string", "a string, the relevant document for the query based on the query text and image content"\}} \\
- "revised\_hard\_negative\_document": the revised hard negative document, \textcolor{blue}{\{"an empty string", "a string, a hard negative document that only appears relevant to the query and the query image content."\}} \\

\textcolor{red}{For the description, please include the following elements:}\\
- General Description: Provide an overall summary of the image, including the primary objects, scene, and notable features.\\
- Object-Level Details: Identify the individual objects in the image, their attributes (e.g., color, size, position), and their relationships to one another.\\
- Contextual Features: Describe the scene or environment, including background details, lighting, and any actions taking place.\\
- Task-specific Brainstorming: Analyze explore how this image could relate to text (e.g., captions, contextual descriptions).\\

\textcolor{red}{Please adhere to the following guidelines:}

- The task must connect the query image and positive image through their content. It must directly indicate the relation without being overly detailed or abstract.\\
- The query (if not empty) should be \textcolor{blue}{\{"extremely long-tail", "long-tail", "common"\}}, \textcolor{blue}{\{"less than 5 words", "5 to 15 words", "at least 10 words"\}}, \textcolor{blue}{\{"clear", "understandable with some effort", "ambiguous"\}}, and diverse in topic.\\
- The query (if not empty) should effectively associate the query image with the positive image. \\
- All documents (if not empty) must be created independent of the query. Avoid copying the query verbatim. It’s acceptable if some parts of the "positive\_document" are not topically related to the query.\\
- All documents (if not empty) should be at least \textcolor{blue}{\{"10", "30", "200", "300"\}} words long.\\
- The "hard\_negative\_document" (if not empty) contains some useful information, but it should be less useful or comprehensive compared to the "positive\_document".\\
- The text of "task\_instruction" should be in English and others fields should be in \textcolor{blue}{\{language\}}.\\
- Do not provide any explanation in any document (if not empty) on why it is relevant or not relevant to the query.\\
- Do not use the word "query" or "document" in the generated content.\\
- Both the query and documents (if not empty) require \textcolor{blue}{\{"high school", "college", "PhD"\}} level education to understand.\\
- \textcolor{red}{When generating the data, please evaluate the following aspects:}\\
  1. Relevance: Are the generated query and documents (if present) tightly linked to their corresponding images? Does the task instruction effectively connect the query image to the positive image?\\
  2. Plausibility: Are the negative examples, including hard negatives, realistic and similar enough to the positive image to challenge the model, while still being definitively incorrect?\\
  3. Clarity: Is the generated task clear and unambiguous, providing sufficient instruction to connect the query image with the positive image, without being overly specific or abstract?\\
  4. Diversity: Does the generated data introduce variation in task instructions, queries, and documents to avoid repetitive patterns in the dataset?\\
- Provide a detailed evaluation of the data based on the above criteria. For each criterion, explain specific flaws or strengths.\\
- Suggest specific revisions to address any identified weaknesses, ensuring the revised data better aligns with the guidelines and task objectives.\\
- Avoid revisions that overly simplify the task or create unrealistic connections between the query and positive image.\\
- Ensure that revised data maintains consistency with the corresponding image content and retrieval task requirements.\\

Your output must always be a JSON object only. Do not explain yourself or output anything else. Be creative!
    
\end{prompt}
\end{figure*}

\section{Data Examples}
\label{appendix: examples}

\begin{figure*}[p]
	\centering
	\includegraphics[width=0.95\textwidth]{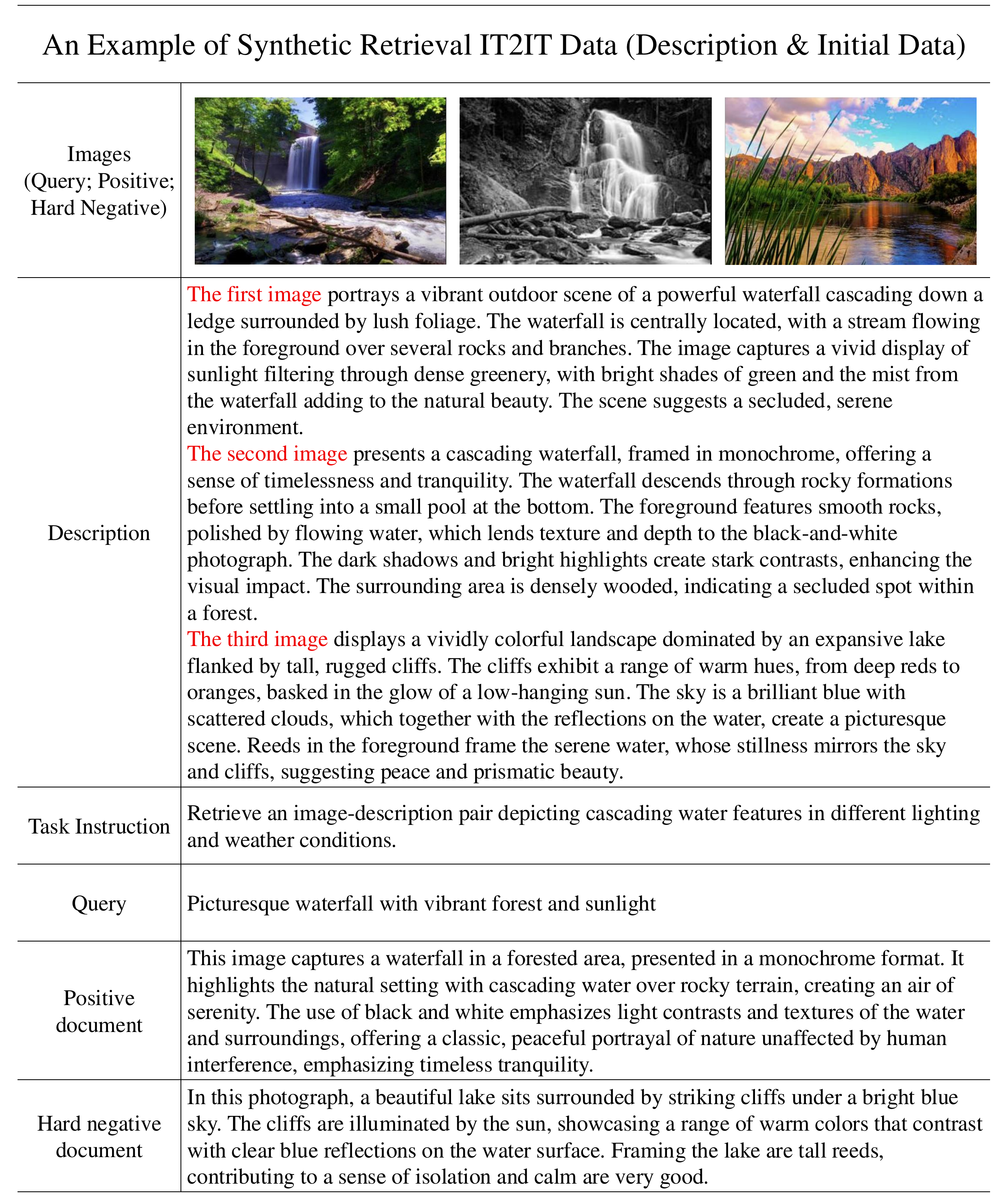}
	\caption{An example of the synthetic Retrieval IT2IT data (part 1). This part includes the input images, the multi-aspect descriptions, and the initially generated data.}
	\vspace{-2ex}
	\label{fig:retrieval_example}
\end{figure*}

\begin{figure*}[p]
	\centering
	\includegraphics[width=0.95\textwidth]{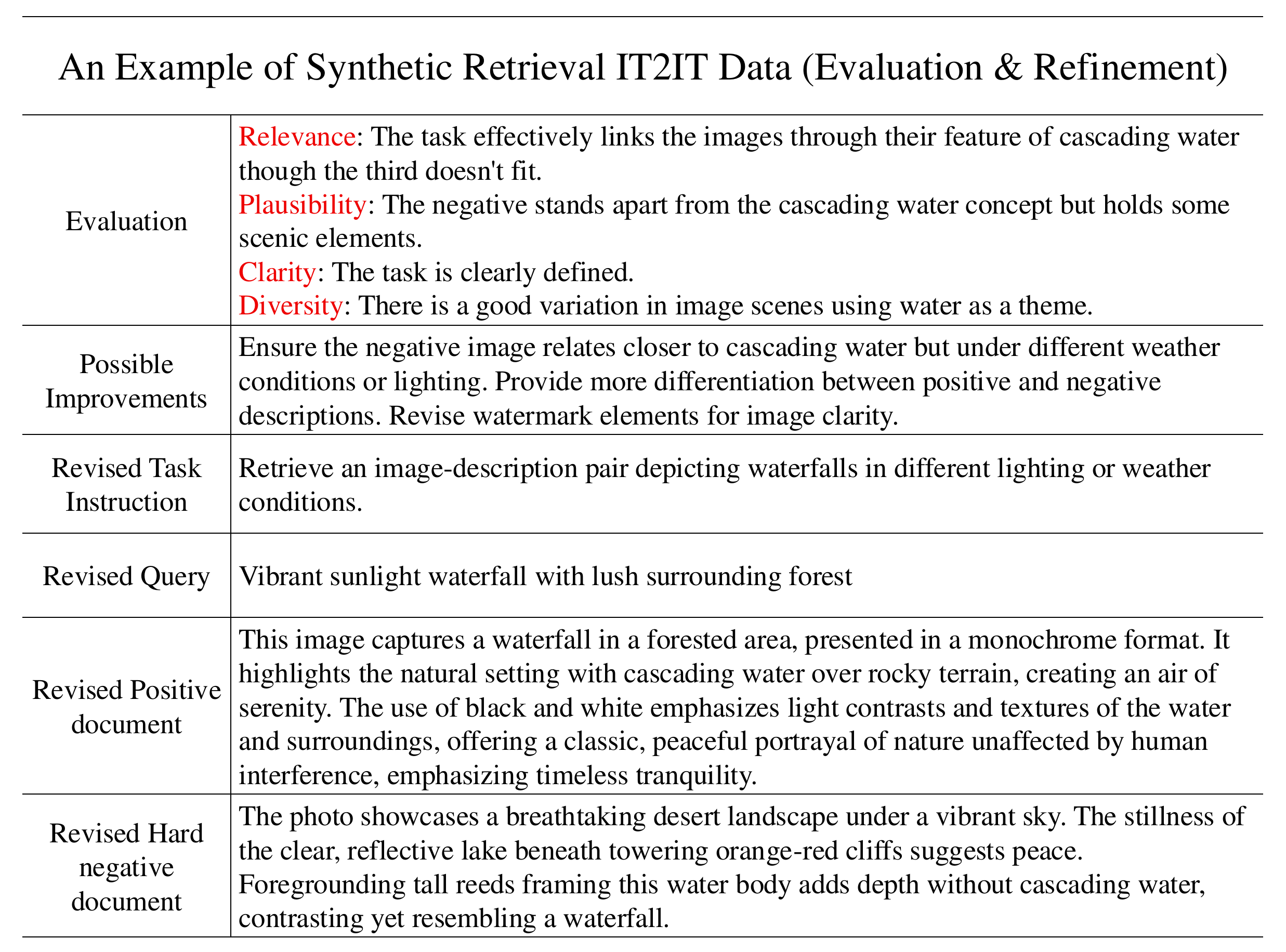}
	\caption{An example of the synthetic Retrieval IT2IT data (part 2). This part includes the evaluation, possible improvements, and the revised data.}
	\vspace{-2ex}
	\label{fig:retrieval_eval_example}
\end{figure*}

\begin{figure*}[p]
	\centering
	\includegraphics[width=0.95\textwidth]{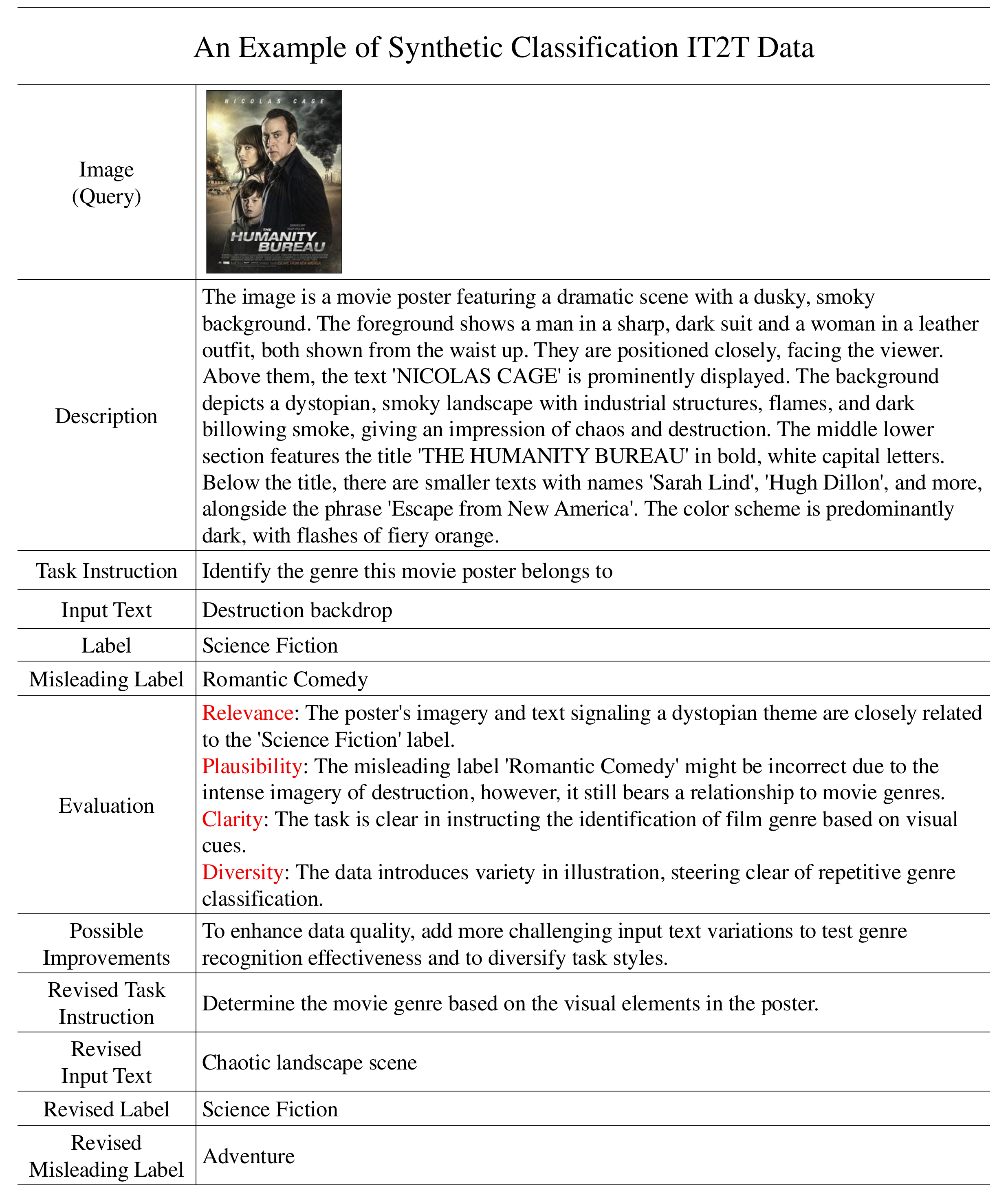}
	\caption{An example of the synthetic Classification IT2T data.}
	\vspace{-2ex}
	\label{fig:classification_example}
\end{figure*}

\begin{figure*}[p]
	\centering 
	\includegraphics[width=0.95\textwidth]{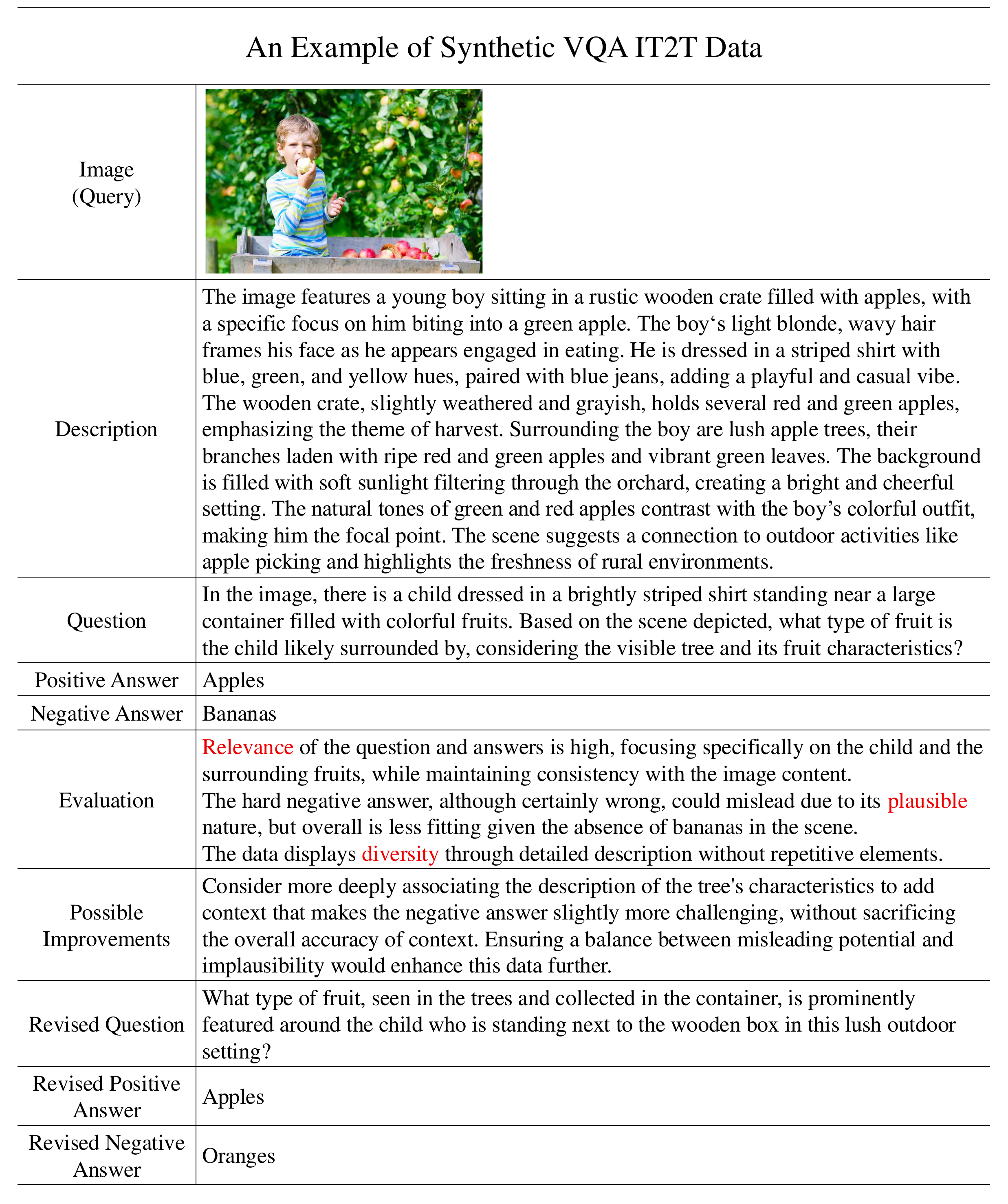}
	\caption{An example of the synthetic VQA IT2T data.}
	\vspace{-2ex}
	\label{fig:vqa_example}
\end{figure*}

In this section, we present the examples of the synthetic multimodal embedding data for Retrieval (Figure~\ref{fig:retrieval_example} and Figure~\ref{fig:retrieval_eval_example}), Classification (Figure~\ref{fig:classification_example}), and VQA (Figure~\ref{fig:vqa_example}) tasks.

\afterpage{\clearpage}

\section{Detailed Results}

In this section, we present the detailed comparisons of \ours{} to baseline models on both zero-shot and supervised finetuning settings on the MMEB benchmark~\cite{MMEB}.
Due to space limitation, we omit the detailed results of partially supervised finetuning models.

\begin{table*}[ht]
\centering
\renewcommand{\arraystretch}{1.2}
\setlength{\tabcolsep}{4pt}
\resizebox{\textwidth}{!}{
\begin{tabular}{lccccccccccccc}
\toprule
\multirow{2}{*}{\textbf{Task}} & \multicolumn{8}{c}{\textit{Zero-shot Setting Models}} & \multicolumn{3}{c}{\textit{Supervised Finetuning Models}}\\ \cmidrule(lr){2-9} \cmidrule(lr){10-12} 
 & \textbf{CLIP} & \textbf{OpenCLIP} & \textbf{SigLIP} & \textbf{BLIP2} & \textbf{MagicLens} & \textbf{E5-V}  & \textbf{MMRet} & \textbf{\ours{}} & \textbf{VLM2Vec} & \textbf{MMRet} & \textbf{\ours{}}\\ 
\midrule
\multicolumn{12}{l}{\textbf{Classification (10 tasks)}} \\
\midrule
ImageNet-1K & 55.8 & 63.5 & 45.4 & 10.3 & 48.0 & 9.6 & 49.1 & 68.8 & 74.5 & 58.8 & 77.8\\
N24News & 34.7 & 38.6 & 13.9 & 36.0 & 33.7 & 23.4 & 45.8 & 54.5 &  80.3 & 71.3 & 81.7\\
HatefulMemes & 51.1 & 51.7 & 47.2 & 49.6 & 49.0 & 49.7 & 51.0 & 55.0 & 67.9 & 53.7 & 64.2\\
VOC2007 & 50.7 & 52.4 & 64.3 & 52.1 & 51.6 & 49.9 & 74.6 & 73.9 & 91.5 & 85.0 & 91.0\\
SUN397 & 43.4 & 68.8 & 39.6 & 34.5 & 57.0 & 33.1 & 60.1 & 72.7 & 75.8 & 70.0 & 77.7\\
\rowcolor[HTML]{FFEB99} Place365 & 28.5 & 37.8 & 20.0 & 21.5 & 31.5 & 8.6 & 35.3 & 39.7 & 44.0 & 43.0 & 43.0\\
\rowcolor[HTML]{FFEB99} ImageNet-A & 25.5 &14.2 &42.6 &3.2 &8.0 &2.0 & 31.6 &46.1 &43.6 & 36.1 & 56.3\\
\rowcolor[HTML]{FFEB99} ImageNet-R & 75.6 & 83.0 & 75.0 & 39.7 & 70.9 & 30.8 & 66.2 & 86.2 & 79.8 & 71.6 & 86.3\\
\rowcolor[HTML]{FFEB99} ObjectNet & 43.4 & 51.4 & 40.3 & 20.6 & 31.6 & 7.5 & 49.2 & 74.8 & 39.6 & 55.8 & 62.5\\
\rowcolor[HTML]{FFEB99} Country-211 & 19.2 & 16.8 & 14.2 & 2.5 & 6.2 & 3.1 & 9.3 & 35.1 & 14.7 & 14.7 & 35.4\\
\rowcolor[HTML]{E8E8E8} \textit{All Classification} & 42.8 & 47.8 & 40.3 & 27.0 & 38.8 & 21.8 & 47.2 & 60.7 & 61.2 & 56.0 & 67.6\\
\midrule
\multicolumn{12}{l}{\textbf{VQA (10 tasks)}} \\
\midrule
OK-VQA & 7.5 & 11.5 & 2.4 & 8.7 & 12.7 & 8.9 & 28.0 & 56.6 & 69.0 & 73.3 & 67.6\\
A-OKVQA & 3.8 & 3.3 & 1.5 & 3.2 & 2.9 & 5.9 & 11.6 & 50.0 & 54.4 & 56.7 & 56.1\\
DocVQA & 4.0 & 5.3 & 4.2 & 2.6 & 3.0 & 1.7 & 12.6 & 81.3 & 52.0 & 78.5 & 90.3\\
InfographicsVQA & 4.6 & 4.6 & 2.7 & 2.0 & 5.9 & 2.3 & 10.6 & 44.0 & 30.7 & 39.3 & 56.5\\
ChartQA & 1.4 & 1.5 & 3.0 & 0.5 & 0.9 & 2.4 & 2.4 & 35.2 & 34.8 & 41.7 & 50.5\\
Visual7W & 4.0 &2.6 &1.2 &1.3 &2.5 &5.8 & 9.0 &40.4 & 49.8& 49.5 & 51.9\\
\rowcolor[HTML]{FFEB99} ScienceQA & 9.4 & 10.2 & 7.9 & 6.8 & 5.2 & 3.6 & 23.3 & 47.3 & 42.1 & 45.2 & 55.8\\
\rowcolor[HTML]{FFEB99} VizWiz & 8.2 & 6.6 & 2.3 & 4.0 & 1.7 & 2.6 & 25.9 & 54.0 & 43.0 & 51.7 & 52.8\\
\rowcolor[HTML]{FFEB99} GQA & 41.3 & 52.5 & 57.5 & 9.7 & 43.5 & 7.8 & 41.3 & 65.4 & 61.2 & 59.0 & 61.7\\
\rowcolor[HTML]{FFEB99} TextVQA & 7.0 & 10.9 & 1.0 & 3.3 & 4.6 & 3.2 & 18.9 & 83.1 & 62.0 & 79.0 & 83.3\\
\rowcolor[HTML]{E8E8E8} \textit{Avg.} & 9.1 & 10.9 & 8.4 & 4.2 & 8.3 & 4.9 & 18.4 & 55.7 &  49.9 & 57.4 & 62.6\\
\midrule
\multicolumn{12}{l}{\textbf{Retrieval (12 tasks)}} \\
\midrule
VisDial & 30.7 & 25.4 & 21.5 & 18.0 & 24.8 & 9.2 & 62.6 & 39.1 &  80.9 & 83.0 & 74.1\\
CIRR & 12.6 & 15.4 & 15.1 & 9.8 & 39.1 & 6.1 & 65.7 & 41.6 &  49.9 &  61.4 & 54.7\\
VisualNews\_t2i & 78.9 & 74.0 & 51.0 & 48.1 & 50.7 & 13.5 & 45.7 & 51.2 &  75.4 & 74.2 & 77.6\\
VisualNews\_i2t & 79.6 & 78.0 & 52.4 & 13.5 & 21.1 & 8.1 & 53.4 & 64.9 & 80.0 & 78.1 & 83.3\\
MSCOCO\_t2i & 59.5 & 63.6 & 58.3 & 53.7 & 54.1 & 20.7 & 68.7 & 55.0 & 75.7 & 78.6 & 76.4\\
MSCOCO\_i2t & 57.7 &62.1 &55.0 &20.3 &40.0 &14.0 & 56.7 &59.1 &73.1 & 72.4 & 73.2\\
NIGHTS & 60.4 & 66.1 & 62.9 & 56.5 & 58.1 & 4.2 & 59.4 & 58.9& 65.5 & 68.3 & 68.3\\
WebQA & 67.5 & 62.1 & 58.1 & 55.4 & 43.0 & 17.7 & 76.3 & 82.9&87.6 & 90.2 & 88.0\\
\rowcolor[HTML]{FFEB99}FashionIQ & 11.4 & 13.8 & 20.1 & 9.3 & 11.2 & 2.8 & 31.5 & 21.6& 16.2 & 54.9 & 28.8\\
\rowcolor[HTML]{FFEB99}Wiki-SS-NQ & 55.0 &44.6 &55.1 &28.7 &18.7 &8.6 & 25.4 & 58.8 &60.2 & 24.9 & 65.8\\
\rowcolor[HTML]{FFEB99}OVEN &  41.1 &45.0 &56.0 &39.5 &1.6 &5.9 & 73.0 &67.6 &56.5 & 87.5 & 77.5\\
\rowcolor[HTML]{FFEB99}EDIS & 81.0 & 77.5 & 23.6 & 54.4 & 62.6 & 26.8 & 59.9 & 55.2 &87.8 & 65.6 & 83.7\\
\rowcolor[HTML]{E8E8E8} \textit{Avg.} & 53.0 & 52.3 & 31.6 & 33.9 & 35.4 & 11.5 & 56.5 & 54.7 &67.4 & 69.9 & 71.0\\
\midrule
\multicolumn{12}{l}{\textbf{Visual Grounding (4 tasks)}} \\
\midrule
MSCOCO & 33.8 & 34.5 & 46.4 & 28.9 & 22.1 & 10.8 & 42.7 & 59.0 &80.6 & 76.8 & 53.7\\
\rowcolor[HTML]{FFEB99}RefCOCO & 56.9 & 54.2 & 70.8 & 47.4 & 22.8 & 11.9 & 69.3 & 78.9 &88.7& 89.8 & 92.7\\
\rowcolor[HTML]{FFEB99}RefCOCO-matching & 61.3 &68.3 &50.8 &59.5 &35.6 &38.9 & 63.2 &80.8 &84.0 & 90.6 & 88.8\\
\rowcolor[HTML]{FFEB99}Visual7W-pointing & 55.1 & 56.3 & 70.1 & 52.0 & 23.4 & 14.3 & 73.5 & 71.2 &90.9 & 77.0 & 92.3\\
\rowcolor[HTML]{E8E8E8}\textit{Avg.} & 51.8 & 53.3 & 59.5 & 47.0 & 26.0 & 19.0 & 62.2 &72.5 &86.1 & 83.6 & 89.6\\
\midrule
\multicolumn{12}{l}{\textbf{Final Score (36 tasks)}} \\
\midrule
\text{All IND Avg.} & 37.1 & 39.3 & 32.3 & 25.3 & 31.0 & 14.9 & 43.5 & 57.2 &  67.5 & 59.1 & 72.3\\
\rowcolor[HTML]{FFEB99}\text{All OOD Avg.} & 38.7 & 40.2 & 38.0 & 25.1 & 23.7 & 11.5 & 44.3 & 60.4 &  57.1 & 68.0 & 66.7\\
\rowcolor[HTML]{E8E8E8}\text{All Avg.} & 37.8 & 39.7 & 34.8 & 25.2 & 27.8 & 13.3 & 44.0 & 58.6 & 62.9 & 64.1 & 69.8\\
\bottomrule
\end{tabular}
}
\caption{Detailed results of zero-shot setting and supervised setting models on each dataset of MMEB~\cite{MMEB}.}
\end{table*}